\documentclass[letterpaper, 10 pt, conference]{ieeeconf}  

\IEEEoverridecommandlockouts                              

\usepackage[pdftex]{graphicx}
\graphicspath{{../pdf/}{../jpeg/}}
\DeclareGraphicsExtensions{.pdf,.jpeg,.png}
\usepackage{algorithm}
\usepackage[tbtags]{amsmath}
\usepackage[noend]{algpseudocode}
\usepackage{euscript}
\usepackage{color}
\usepackage{booktabs} 

\makeatletter
\let\NAT@parse\undefined
\makeatother
\usepackage[numbers,sort,compress]{natbib}
\usepackage{amsfonts}
\usepackage{amssymb}
\usepackage{mathtools}
\usepackage{microtype}
\usepackage{soul}
\usepackage{tikz} 
\usepackage{bm}

\usepackage[english]{babel}
\usepackage[scr=boondox]{mathalfa}

\usepackage{xparse}
\ExplSyntaxOn
\cs_new_eq:NN \calc \fp_eval:n
\ExplSyntaxOff

\usepackage[caption=false,font=footnotesize]{subfig}

\usepackage[hidelinks]{hyperref}
\usepackage[capitalise]{cleveref}

\crefformat{section}{#2Sec.~{\textup{#1}}#3}
\crefformat{figure}{#2Fig.~{\textup{#1}}#3}
\crefformat{equation}{#2({\textup{#1}})#3}
\crefmultiformat{figure}{Figs.~#2#1#3}{ and~#2#1#3}{}{}
\crefmultiformat{section}{Secs.~#2#1#3}{ and~#2#1#3}{}{}
\crefrangeformat{section}{Secs.~#3#1#4 to~#5#2#6}

\newcommand{\transform}[1][]{%
	\ifthenelse{\equal{#1}{}}{\mathbf{T}}{\mathbf{T}_{#1}}%
}

\newcommand{\mylabel}{\ell}
\newcommand{\se}[1][3]{$SE\left(#1\right)$}
\newcommand{\kjtodo}[1]{{{{#1}}}}

\makeatletter
\def\subsubsection{\@startsection{subsubsection}{3}{\z@}{0ex plus 0.1ex minus 0.1ex}{0ex}{\normalfont\normalsize\itshape}}
\makeatother

\newcommand{\disctraj}{\mathscr{S}}
\newcommand{\state}{\mathbf{x}}

\title{\LARGE \bf
Occlusion-Robust MVO:\\
Multimotion Estimation Through Occlusion Via Motion Closure
}

\author{Kevin M. Judd$^{1}$ and Jonathan D. Gammell$^{1}$
\thanks{$^{1}$K. M. Judd and J. D. Gammell are with the Estimation, Search, and Planning (ESP) research group at the Oxford Robotics Institute (ORI), University of Oxford, United Kingdom. {\tt\footnotesize \{kjudd, gammell\}@robots.ox.ac.uk}}%
}

\begin{document}

\maketitle
\thispagestyle{empty}
\pagestyle{empty}

\begin{abstract}
Visual motion estimation is an integral and well-studied challenge in autonomous navigation.
Recent work has focused on addressing multimotion estimation, which is especially challenging in highly dynamic environments.
Such environments not only comprise multiple, complex motions but also tend to exhibit significant occlusion.

Previous work in object tracking focuses on maintaining the integrity of object tracks but usually relies on specific appearance-based descriptors or constrained motion models.
These approaches are very effective in specific applications but do not generalize to the full multimotion estimation problem.\looseness=-1

This paper presents a pipeline for estimating multiple motions, including the camera egomotion, in the presence of occlusions.
This approach uses an expressive motion~prior to estimate the \se{} trajectory of every motion in the scene, even during temporary occlusions, and identify the reappearance of motions through \emph{motion closure}. The performance of this occlusion-robust multimotion visual odometry (MVO) pipeline is evaluated on real-world data and the Oxford Multimotion Dataset.\looseness=-1
\end{abstract}

\section{Introduction} \label{sec:introduction}
The ability to safely navigate through a dynamic environment is a crucial task in autonomous robotics.
Visual odometry (VO) is widely used to estimate the egomotion of a camera by isolating the static parts of a scene \cite{moravec1980}, but it is more challenging to segment multiple motions in a complex, dynamic scene.
Recent work has focused on extending  VO approaches to this \emph{multimotion estimation problem} \cite{judd2018}.
These highly dynamic scenes not only pose difficult motion estimation challenges but also tend to include significant amounts of occlusion.\looseness=-1

An occlusion can be defined as any lack of direct observations of a scene, regardless of the cause.
Direct occlusions occur when an object is obscured by another or leaves the sensor's field of view. 
Indirect occlusions result from sensor limitations or algorithmic failure, such as when motion blur 
corrupts feature matching or object detection.
Consistently estimating multiple, continuous motions in the presence of both direct and indirect occlusions is necessary for autonomous navigation in complex, dynamic environments.

\begin{figure}[t!]
	\vspace{1.8mm}
	\centering
	
	\subfloat[]{\label{fig:marquee:before}\includegraphics[width=.98\columnwidth]{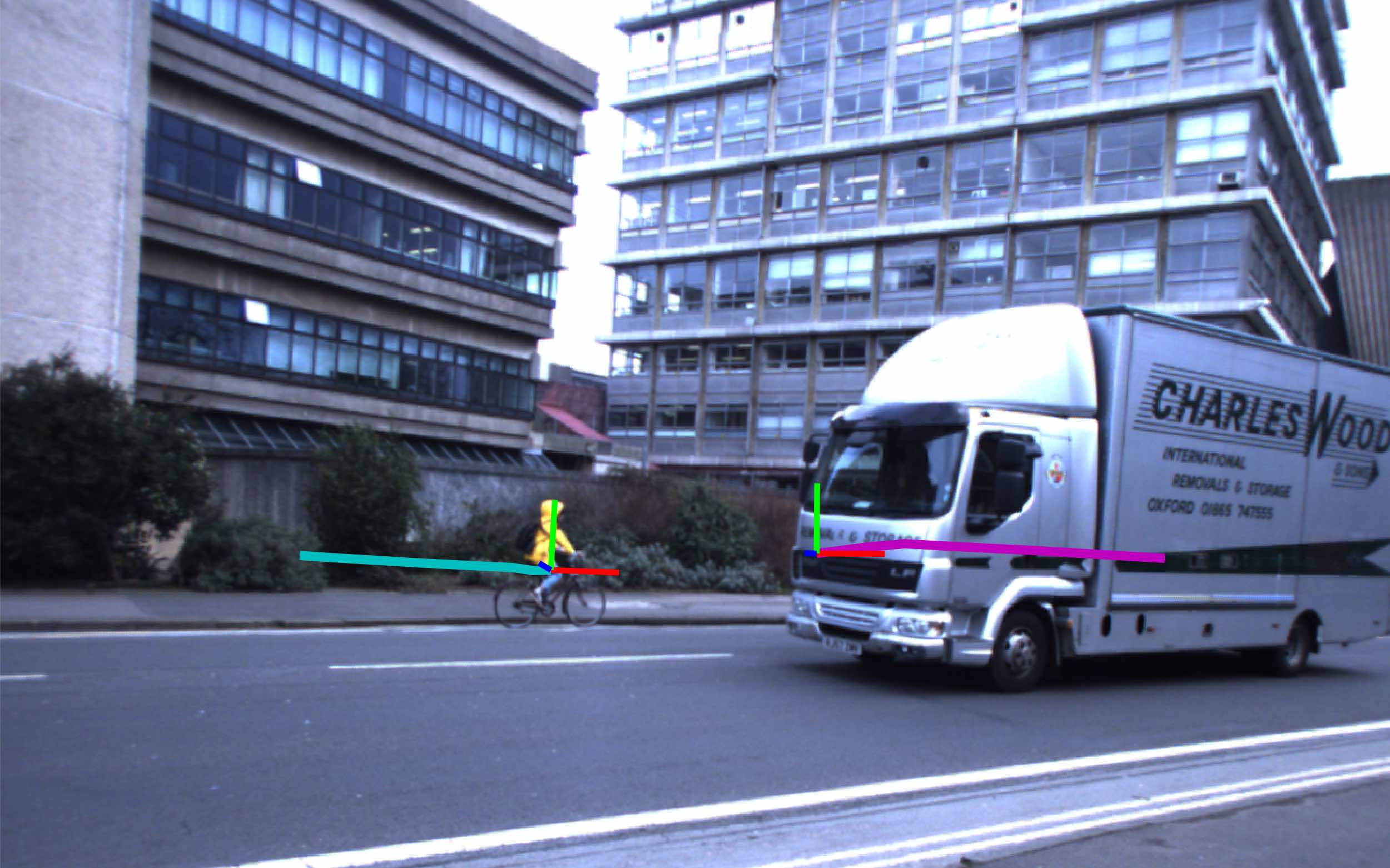}}
	\vspace{-2mm}
	\subfloat[]{\label{fig:marquee:after}\includegraphics[width=.98\columnwidth]{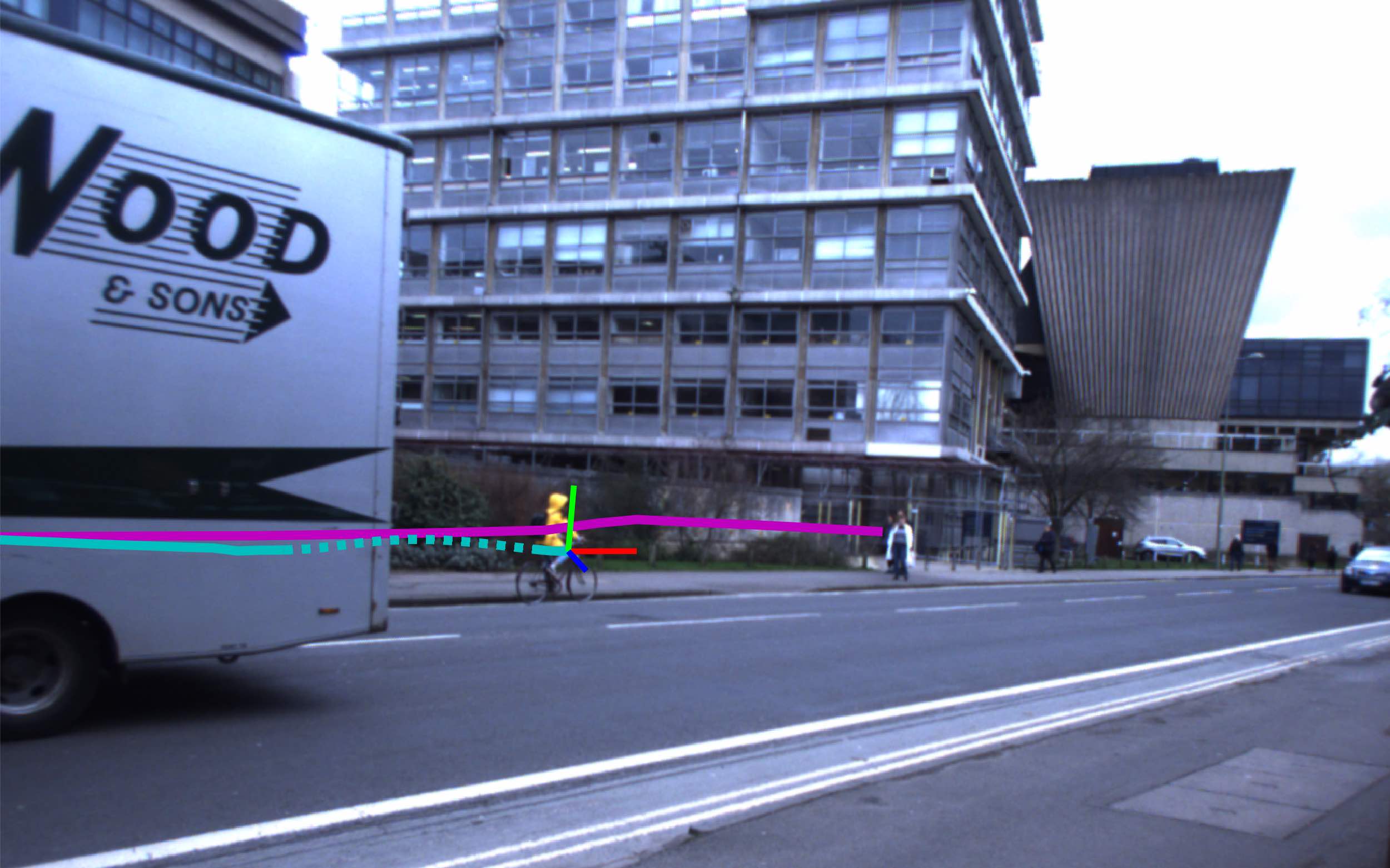}}
	\vspace{-1mm}

	\caption{Motion trajectories produced by the occlusion-robust MVO system for a real-world street scene before (a) and after (b) an occlusion.
		The camera pans to follow a cyclist (cyan) moving from left to right across the scene as they are temporarily occluded by a truck (purple) moving from right to left.
		The egomotion of the camera is estimated simultaneously with the trajectories of the cyclist and truck.
		The object trajectories are estimated directly when they are visible (solid lines), and a white-noise-on-acceleration prior is used to extrapolate the trajectory of the cyclist while it is occluded by the truck (dashed line).
		When the cyclist becomes unoccluded, it is detected through \emph{motion closure} and its extrapolated trajectory is rectified through interpolation.
		The result is a consistent \se{} trajectory of every motion in the scene, even during temporary occlusions.\looseness=-1}
		\vspace{-4.75mm}
	\label{fig:marquee}
\end{figure}

Multiple object tracking (MOT) approaches address occlusions using appearance-based object representations to recognize objects when they become unoccluded \cite{yilmaz2006}.
These approaches employ specialized detectors and simple motion models that limit their ability to track general objects and estimate the full \se{} pose of each object.

Our multimotion visual odometry (MVO) pipeline \cite{judd2018} simultaneously estimates the full \se{} trajectory of every motion in a scene, including the  camera egomotion, without \emph{a priori} assumptions about object appearance.
It addresses the multimotion estimation problem by applying multilabeling techniques to the traditional VO pipeline using only a rigid-motion assumption.
By emphasizing motion over appearance, we contend that it is generally more important to understand \emph{how} things are moving in the environment than to understand \emph{what} they are.

\begin{figure*}[t]
	\centering{
		\subfloat[]{\label{main:a}\includegraphics[width=.305\textwidth]{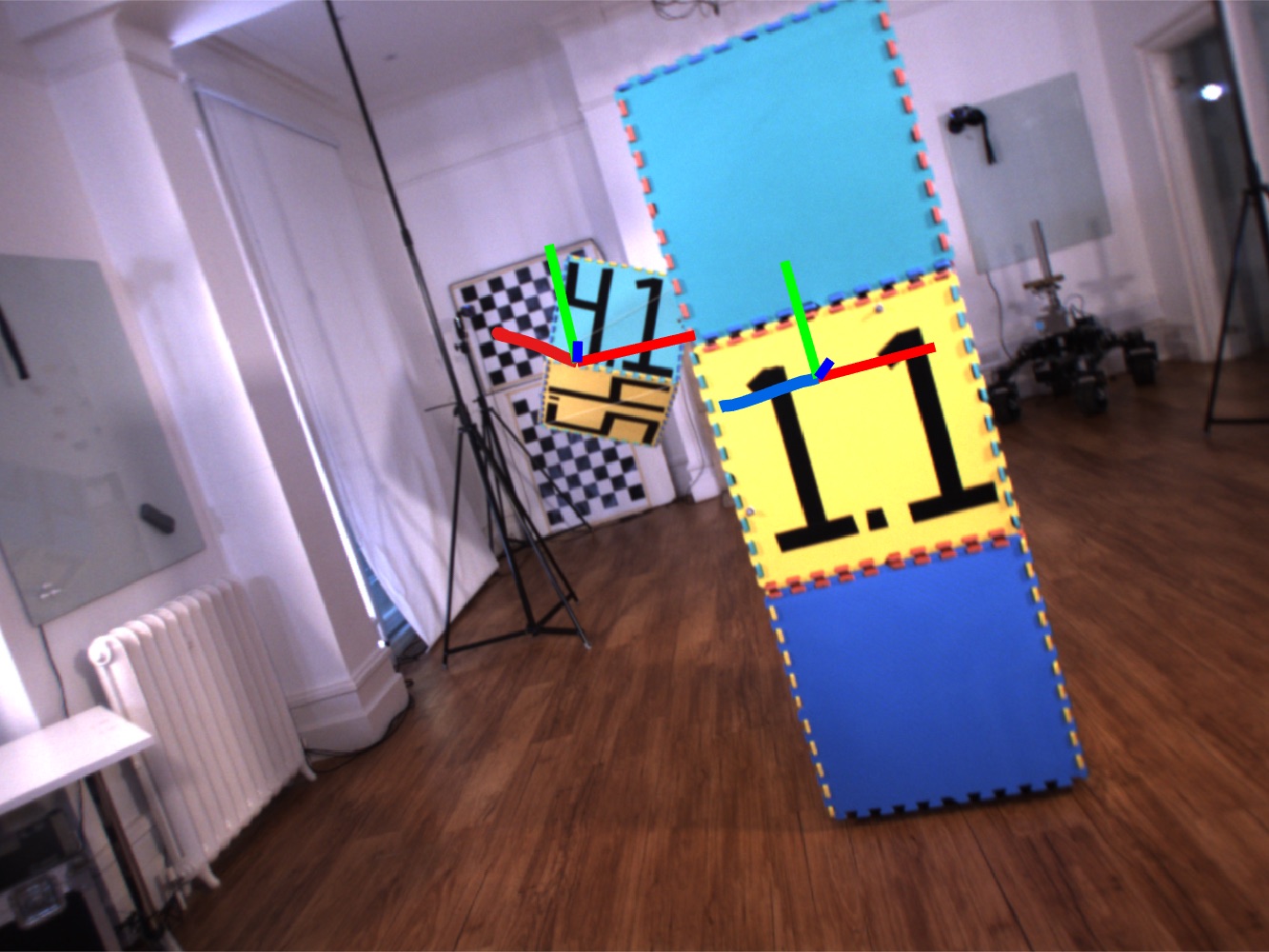}}\hfill
		\subfloat[]{\label{main:b}\includegraphics[width=.305\textwidth]{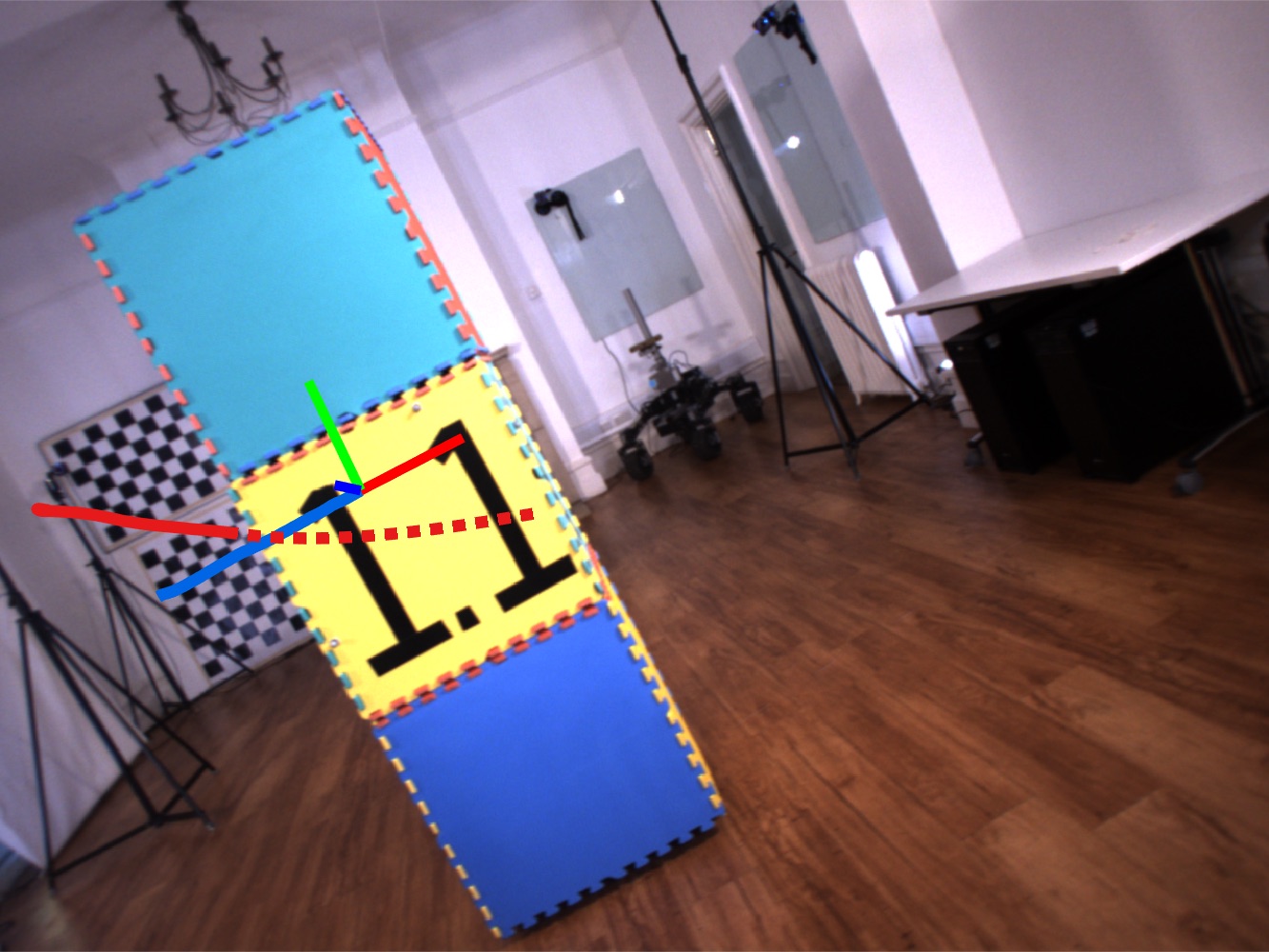}}\hfill
		\subfloat[]{\label{main:c}\includegraphics[width=.305\textwidth]{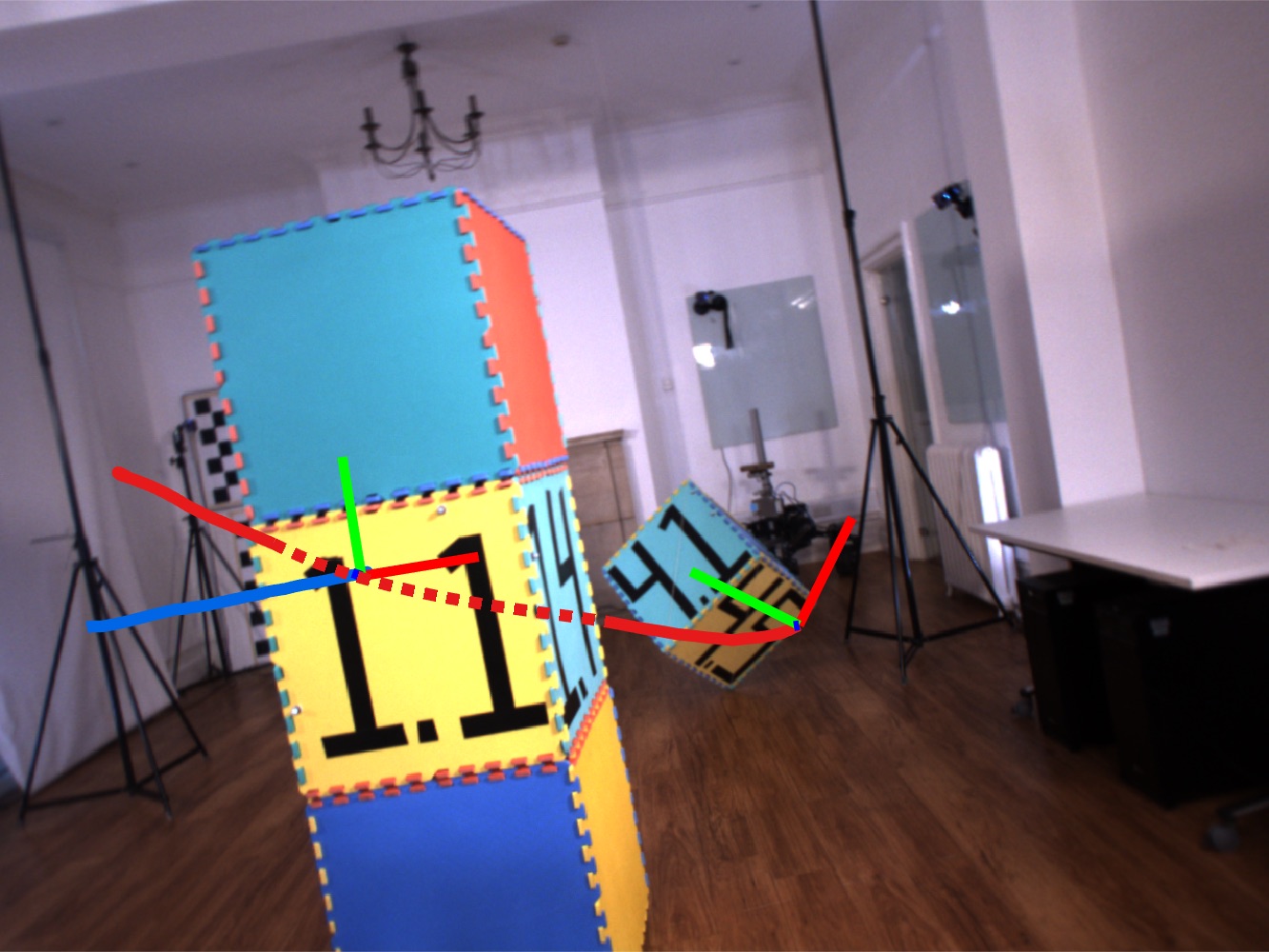}}}
	
	\caption{Trajectory estimates from the occlusion-robust MVO system before (a), during (b), and after (c) an occlusion in the \texttt{occlusion\_2\_unconstrained} segment of the OMD \cite{judd2019}. 
		The trajectory of the swinging block (4, red) is directly estimated when it is visible in (a) and (c) and is extrapolated using the constant-velocity motion prior (dashed line) when it is occluded by the block tower (1, blue) in (b).
		When the block becomes unoccluded in (c), it is rediscovered through \emph{motion closure} and the estimates are interpolated to match the directly estimated trajectory.
		\looseness=-1
	}
	\label{fig:occlusion}
\end{figure*}

This paper presents the occlusion-robust MVO pipeline, which extends the MVO framework to estimate multiple \se{} motions through occlusion (\cref{fig:marquee}).
It exploits a continuous \se{} motion prior for egomotion estimation and extends it to geocentric, third-party motion estimation.
This prior allows occlusion-robust MVO to both extrapolate temporarily occluded motions and recognize when they reappear (\cref{fig:occlusion}).
This process of \emph{motion closure} improves trajectory consistency and accuracy without relying on specific appearance-based models.

The accuracy of this approach is evaluated both quantitatively and qualitatively. 
The metric estimation error is measured against ground-truth data from the Oxford Multimotion Dataset (OMD) \cite{judd2019}.
The applicability of both the rigid-motion and constant-velocity assumptions is demonstrated on a real-world scenario with a truck occluding a cyclist (i.e., a rigid motion occluding a piecewise-rigid motion).

\section{Background} \label{sec:background}
Motion estimation and object tracking are integral to a wide range of computer vision applications.
Tracking and estimating the motion of an individual object through a scene has been widely explored but complex, dynamic scenes make this task significantly more difficult due to frequent occlusions.
Successfully addressing the multimotion estimation problem requires the ability to both track and estimate multiple motions in the presence of direct and indirect occlusions.\looseness=-1

\subsection{Multimotion Estimation}
Many multimotion estimation approaches only solve a subset of the rigid multiomotion estimation problem by applying application-specific constraints and assumptions.
This limits their applicability for real-world multimotion estimation problems.

Costeira and Kanade \cite{costeira1998} use matrix decomposition to determine the motion and shape of each dynamic object.
This factorization usually requires points to be tracked for the entire estimation window, which is difficult due to direct and indirect occlusions.
Some techniques allow for missing data points \cite{vidal2004} but are not designed for many short feature tracks, as is commonly encountered in practice. \looseness=-1

Torr \cite{torr1998} uses a recursive RANSAC framework to find and remove dominant motion models from the remaining feature points.
This framework is efficient at finding the dominant models in a scene, but the likelihood of sampling consistent models decreases as models are removed and the signal-to-noise ratio of the remaining points decreases.
Sabzevari and Scaramuzza \cite{sabzevari2016} improve this likelihood by applying geometric and kinematic constraints specific to driving scenarios, but these constraints do not generalize well to other applications.\looseness=-1

Ozden et al. \cite{ozden2010} consider many practical challenges in multimotion estimation, such as incomplete feature tracks.
They propose a model selection framework that explicitly models the merging and splitting of motions, but it relies on separate egomotion estimation and does not address direct occlusions.\looseness=-1

Qiu et al. \cite{qiu2019} and Eckenhoff et al. \cite{eckenhoff2019} attempt to address the multimotion estimation problem using monocular visual-inertial odometry techniques.
Monocular cameras cannot directly estimate depth, so an IMU must be used to disambiguate the scale of the scene.
These limitations lead to several degenerate cases in multimotion estimation and these techniques do not address occlusion.

The original MVO pipeline \cite{judd2018} addresses the multimotion estimation problem by applying multilabeling techniques to the traditional VO pipeline.
MVO simultaneously estimates the full \se{} trajectory of the camera and every third-party motion in a scene without \emph{a priori} assumptions about object appearance.
It relies on direct observations and can estimate through some partial and indirect occlusions, but it is unable to handle significant observation dropouts.

\subsection{Multiple Object Tracking}
Most visual MOT techniques follow the \emph{tracking-by-detection} paradigm.
They use a variety of specific, appearance-based object models to detect targets in each frame and focus on accurately associating those detections across multiple frames \cite{reid1979}.\looseness=-1

Target detectors often use bounding-box representations rather than the full target pose, so objects are usually tracked in image or Cartesian space using simple motion models {\cite{yilmaz2006}}.
Other approaches define highly specialized models that do not generalize well to other domains {\cite{wu2007,shu2012,yang2012,mitzel2010}}.
These constraints limit their ability to track general objects and estimate the full \se{} pose of each object.
\enlargethispage*{\baselineskip}
\pagebreak

Tracking and data association are often performed using recursive filters or global energy minimizations.
Kalman \cite{reid1979,khan2004} and particle \cite{breitenstein2009} filters use simple motion models to recursively predict the location of the target and update the current state based on current observations.
The problem of assigning tracks to new detections or other tracks can be solved using the Hungarian algorithm \cite{xing2009} or other greedy alternatives \cite{breitenstein2009, wu2007}.
Tracking-by-detection techniques are limited by the quality of the detectors they use, and recursive methods often fail due to occlusions and lighting changes.\looseness=-1

\begin{figure*}[t]
	\centering%
	\includegraphics[clip,width=.99\textwidth]{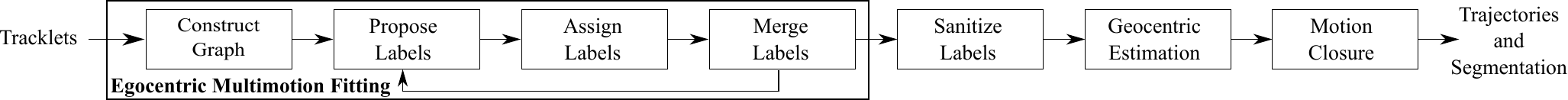}
	\caption{Occlusion-aware MVO extends the original MVO pipeline to accurately estimate trajectories through occlusions. 
		The multimotion-fitting section of the pipeline both segments a set of tracklets according to their motion and estimates the trajectories explaining that motion.
		Once the segmentation converges, the camera egomotion is identified and used to estimate the geocentric trajectories of all other objects in the scene.
		In \emph{motion closure}, the white-noise-on-acceleration prior is used to extrapolate occluded trajectories and determine if newly discovered motions can be explained by the reappearance of an occluded object.\looseness=-1}
	\vspace{-4mm}
	\label{fig:pipeline}
\end{figure*}

Energy-based techniques incorporate object appearance, motion, and interaction models in a cost functional.
The functional is defined over a graph where vertices represent detections and edges represent transitions between frames, and it is minimized using flow-based techniques \cite{zhang2008} or optimization frameworks \cite{byeon2018}.
These specialized approaches must define representative cost functionals and do not generalize well to other applications.

\subsection{Tracking Through Occlusion}
Accurate data association is more difficult in highly dynamic environments with significant occlusion.
Direct occlusions can be predicted by modeling object overlaps \cite{mitzel2010} or using scene understanding \cite{kaucic2005}, which can help to avoid misassociated detections.
Yang et al. \cite{yang2011} propose a learning-based conditional random field model that considers the interdependence of observed motions, especially in the presence of occlusion.
These prediction methods can be used for direct occlusions, but indirect occlusions are more difficult to predict.\looseness=-1

Partial occlusions are challenging for appearance-based techniques because they change the observed shape of the occluded object.
Feature-based techniques track targets through partial occlusions when a sufficient number of feature points can be tracked \cite{sugimura2009}, but grouping features into distinct objects is difficult if their bulk motion is similar.
Other techniques define specific, part-based appearance models to infer the pose of the entire object from the portions that are visible \cite{wu2007,hu2012,shu2012}.\looseness=-1

Full occlusions are often overcome by using motion priors to extrapolate trajectories in the absence of direct observations. 
Zhang et al. \cite{zhang2008} explicitly incorporate occlusion hypotheses into a flow minimization.
This \emph{hypothesize-and-test} paradigm works well in the presence of short or partial occlusions but fails under long occlusions as there is no information available to prune hypotheses \cite{reid1979}. 
Ryoo et al. \cite{ryoo2008} avoid this impractical growth with their \emph{observe-and-explain} paradigm, which delays hypothesizing occluded motions until an unoccluded detection is observed near the source of occlusion.
Mitzel et al. \cite{mitzel2010} extrapolate unobserved target trajectories for a set number of frames to allow for reassociation when the target becomes unoccluded.
These techniques are limited by the object representations used by the target detectors, so they tend to be constrained to simple motion models that do not adequately represent the motion of an object.
They also focus on rediscovering an object based on appearance and position, tending to ignore its current motion.\looseness=-1

\vspace{10ex}

This paper extends our previous work in multimotion estimation \cite{judd2018} to estimate and track motions through occlusions.
It introduces a continuous motion prior that is used to extrapolate motions through both direct and indirect occlusions and allows for \emph{motion closure} to reacquire motions and improve their occluded estimates.
The full \se{} trajectory of every motion in the scene is estimated through occlusions, and the approach is evaluated on real-world data and an occlusion dataset containing ground-truth trajectories for all motions in the scene.\looseness=-1


\section{Motion Priors} \label{sec:priors}
The way a motion is modeled directly affects the performance of multimotion estimation.
Overly-simplified models can reduce estimation accuracy, but complex models are often inefficient. 
A good compromise is rigid motion, which reduces the motion trajectory space to \se{} while maintaining fidelity with most real-world motions.
Many~approaches sacrifice this fidelity for simplicity by further constraining motions to \se[2]{} \cite{sabzevari2016, mitzel2010}, $\mathbb{R}^3$ \cite{byeon2018,zhang2008}, or image space \cite{ryoo2008}.
Others sacrifice generality by defining high-dimensional models for applications such as human tracking \cite{wu2007,shu2012,yang2012}.

Models can also be defined discretely or continuously.
Discrete models represent a trajectory as a sparse set of states, which is well-suited for synchronized sensors such as globally shuttered cameras.
Continuous models smoothly represent a trajectory at all times using an assumption, or \emph{prior}, about how objects move through the world \cite{anderson2015}.
These models incorporate this smooth prior directly into the representation, which is preferable for scanning or high-rate sensors.
This prior can also be exploited to intelligently estimate occluded trajectories.

Motions can also be defined in arbitrary frames, and a simple motion in one frame may become complex when expressed in another.
Two bodies, each with some known motion relative to a static reference frame, do not exhibit the same type of motion relative to \emph{each other}.
A model that is expressed egocentrically may be appropriate to estimate the egomotion of a camera relative to its static environment but not relative to other dynamic objects.
It is therefore often necessary to express models in an inertial or quasi-inertial (e.g., geocentric) frame.\looseness=-1

\subsection{White-Noise-on-Acceleration Motion Prior}
This paper extends the \se{} white-noise-on-acceleration (i.e., locally-constant-velocity) motion prior described by Anderson et al. \cite{anderson2015} to all motions in the scene.
This prior effectively penalizes the trajectory's deviation from a constant body-centric velocity. 
It is physically founded because objects tend to move smoothly (i.e., with differentiable velocity) and drastic changes in direction, especially those that coincide with occlusions, are relatively rare in many autonomous navigation scenarios.

The continuous-time trajectory of the $\ell$-th motion in a scene, $\disctraj{}_{\mylabel{}}\left(t\right)\coloneqq\left\{\transform[\mylabel{}]\left(t\right), \boldsymbol{\varpi}_\mylabel{}\left(t\right)\right\}$, is defined as both the \se[3]{} poses, $\transform[\mylabel{}]\left(t\right)$, and the local, body-centric velocities, $\boldsymbol{\varpi}_\mylabel{}\left(t\right)$.
The trajectory state is assumed to vary smoothly over time in the Lie algebra, $\mathfrak{se}\left(3\right)$.
This prior takes the form
\begin{equation}
\begin{aligned}
\dot{\mathbf{T}}_{\mylabel{}}\left(t\right) &= {\boldsymbol{\varpi}}_{\mylabel{}}\left(t\right)^{\wedge} {\transform[\mylabel{}]\left(t\right)}\\
\dot{\boldsymbol{\varpi}}_{\mylabel{}}\left(t\right) &= \mathrm{\textbf{w}}'\left(t\right), \qquad \mathrm{\textbf{w}}'\left(t\right) \sim \mathcal{GP}\left(\mathbf{0},\mathbf{Q}_c'\delta\left(t-t'\right)\right),
\end{aligned}
\label{eq:nonlinear}
\end{equation}
where $\mathrm{\textbf{w}}'$ is a zero-mean, white-noise Gaussian process with power spectral density matrix, $\mathbf{Q}_c' \in \mathbb{R}^{6\times6}$, and $\boldsymbol{\varpi}^{\wedge}$ is the $\mathfrak{se}\left(3\right)$ representation of $\boldsymbol{\varpi} \in \mathbb{R}^6$ as defined in \cite{barfoot2017}.

This continuous-time trajectory can be estimated at a collection of discrete time steps, $t_1, \dots, t_K$, such that,
\begin{equation*}
\disctraj{}_{\mylabel{}_{k}} \coloneqq \left\{\transform[\mylabel{}_{k}\mylabel{}_{1}], \boldsymbol{\varpi}_{\mylabel{}_k} \right\} \equiv \disctraj_{\ell}\left(t_k\right), \qquad t_1 \leq t_k \leq t_K,
\end{equation*}
where $\transform[\mylabel{}_{k}\mylabel{}_{1}] \coloneqq \transform[\mylabel{}]\left(t_k\right)$ and $\boldsymbol{\varpi}_{\mylabel{}_k} \coloneqq \boldsymbol{\varpi}_\mylabel{}\left(t_k\right)$. 
These time steps correspond to observation times when measurements of the scene are collected.

The system in \eqref{eq:nonlinear} is nonlinear and finding a numerical solution is costly. 
By assuming the motion between measurement times is small, the system can be recast as a set of local, linear time-invariant stochastic differential equations.
Under the constant-velocity assumption with no exogenous input, these equations take the form, 
\begin{equation}
\hspace{-0.75mm}\boldsymbol{\gamma}_{\mylabel{}_k}\left(\tau\right) = \boldsymbol{\Phi}\left(\tau,t_k\right)\boldsymbol{\gamma}_{\mylabel{}_k}\left(t_k\right) = \begin{bmatrix}
\mathbf{1} & \hspace{-1.85mm}\left(\tau-t_k\right)\mathbf{1} \\
\mathbf{0} & \mathbf{1}
\end{bmatrix}\boldsymbol{\gamma}_{\mylabel{}_k}\left(t_k\right),
\label{eq:transition}
\end{equation}
where $\boldsymbol{\Phi}\left(\tau,t_k\right)$ is the state transition function from $t_k$ to $\tau$ and $\mathbf{1}$ is the identity matrix. The local state, $\boldsymbol{\gamma}_{\mylabel{}_k}$, is defined as\looseness=-1
\begin{equation*}
\boldsymbol{\gamma}_{\mylabel{}_k}\left(t\right) \coloneqq \begin{bmatrix}
\ln\left(\transform[\mylabel{}]\left(t\right){\transform[{\mylabel{}_k}{\mylabel{}_1}]^{-1}}\right)^{\vee}\\
{\boldsymbol{\mathcal{J}}\left(\ln\left(\transform[\mylabel{}]\left(t\right) {\transform[{\mylabel{}_k}{\mylabel{}_1}]}^{-1}\right)^{\vee}\right)^{-1} {\boldsymbol{\varpi}_{\mylabel{}}\left(t\right)}}
\end{bmatrix}, 
\label{eq:localstate}
\end{equation*}
where $\boldsymbol{\mathcal{J}}\left(\cdot\right)$ is the left Jacobian of \se[3]{}, and $t_k \leq t \leq t_{k+1}$. 
Applying the prior \emph{locally} at each time, $k$, casts the \emph{global} nonlinear system as a sequence of linear, time-invariant systems.\looseness=-1

\section{Methodology} \label{sec:methodology}
The original MVO pipeline \cite{judd2018} extends VO to \emph{multimotion} segmentation and estimation. 
As with traditional stereo VO pipelines, a set of tracklets $\mathcal{P} \coloneqq \left\{p\right\}$ is generated by matching salient image points across rectified stereo image pairs and temporally across consecutive stereo frames. 
The motion segmentation and estimation are then cast as a multilabeling problem where each label, $\mylabel{}\in\mathcal{L}$, represents a motion hypothesis, $\disctraj{}_\mylabel{}$, calculated from a subset of tracklets, $\mathcal{P}_\mylabel{} \subseteq \mathcal{P}$.
The labeling is found using CORAL \cite{amayo2018}, a convex optimization approach to the multilabeling problem.
All motion hypotheses are initially treated as potentially belonging to the static portions of the scene (i.e., represent the camera's egomotion).
Geocentric trajectories are found in a final step where a label is selected to represent the motion of the camera.\looseness=-1

The occlusion-robust MVO pipeline (\cref{fig:pipeline}) extends MVO to handle both direct and indirect occlusions by employing a continuous \se{} motion prior.
The prior is used both to estimate directly observed trajectories and to extrapolate occluded motions.
All motion hypotheses are treated as potentially describing the egomotion until the segmentation converges (\cref{sec:methodology:egocentric}).
The egomotion label is identified before performing a full-batch estimation of each trajectory in a geocentric frame (\cref{sec:methodology:geocentric}). 
The motion prior is used to extrapolate previously estimated trajectories that are not found in the current frame due to occlusion or estimation failure (\cref{sec:methodology:extrapolation}).
These extrapolated trajectories are then used in \emph{motion closure} to determine if any new trajectory can be explained by the reappearance of a previously observed motion.
Trajectories found to belong to the same motion are used to correct occluded estimates through interpolation (\cref{sec:methodology:motionclosure}).\looseness=-1




\subsection{Continuous-Time Geocentric Estimation}\label{sec:methodology:egocentric}
The original MVO pipeline estimates motion trajectories without assuming which motion represents the camera \cite{judd2018}, but this is not appropriate for the white-noise-on-acceleration prior.
This locally-constant-velocity assumption is not generally valid for two noninertial frames because two bodies moving with constant velocity relative to an inertial frame do not generally have zero acceleration relative to \emph{each other} (\cref{sec:priors}). 
The camera egomotion is estimated from the static background but estimates of third-party motions must account for camera motion, which may result in nonconstant relative velocity.
Addressing this requires identifying and estimating the camera egomotion first before using it to estimate the other trajectories in a geometric frame.

The egomotion label, $C$, is chosen using prior information or heuristics.
It can be initialized as the largest label, as in VO, 
after which it can be propagated forward in time by choosing the label that maximizes the overlap in support with the previous egomotion  label.
Motion-based similarity metrics can also be used to identify or validate the choice at each timestep and maintain a consistent egomotion trajectory.

\subsubsection{Egomotion Estimation}
The egomotion of the camera is estimated using the approach described in \cite{anderson2015}.
The system state, $\state{}$, comprises the estimated pose transforms and body-centric velocities, $\left\{\disctraj{}_{C_k}\right\}_{k=1,\dots, K}$, and the labeled landmark points, $\left\{\mathbf{p}_{C_1}^{{j_1}{C_1}}\right\}_{j=1,\dots,\vert\mathcal{P}_C\vert}$. 
The estimated state, $\state{}$, is found by minimizing an objective function, $J\left(\state{}\right) = J_{y}\left(\state{}\right) + J_{p}\left(\state{}\right)$, consisting of the measurement and prior terms.\looseness=-1


The measurement term, $J_y\left(\state{}\right)$, constrains the estimated state using the observations,
\begin{equation*}
J_{y}\left(\state{}\right) \coloneqq \frac{1}{2}\sum_{jk}\mathbf{e}_{y,jk}\left(\state{}\right)^{T}\mathbf{R}_{jk}^{-1}\mathbf{e}_{y,jk}\left(\state{}\right),
\end{equation*}
where the error is the residual of the measurement model, $\mathbf{g}\left(\cdot\right)$, compared to the observations, $\mathbf{y}_{jk}$.
\begin{equation}
\mathbf{e}_{y,jk}\left(\state{}\right) \coloneqq \mathbf{y}_{jk} - \mathbf{g}\left(\state{}_{jk}\right) = \mathbf{y}_{jk} - \mathbf{s}\left(\mathbf{z}\left(\mathbf{x}_{jk}\right)\right).
\label{eq:meas_error}
\end{equation}
and $\state{}_{jk} \coloneqq \left\{\disctraj{}_{C_k},\mathbf{p}_{C_1}^{{j_1}{C_1}}\right\}$ is the estimation state. 
The measurement model applies the perspective camera model, $\mathbf{s}\left(\cdot\right)$,  to landmark points transformed by the transform model, $\mathbf{z}\left(\mathbf{x}_{jk}\right) \coloneqq {\transform[{C_{k}C_{1}}]}\mathbf{p}_{C_1}^{{j_1}{C_1}}$, and $\mathbf{R}_{jk}$ is the covariance of the zero-mean additive Gaussian noise in the measurements.

The prior term, $J_p\left(\state{}\right)$, constrains the current trajectory estimate by the previous velocity, 
\begin{equation*}
J_{p}\left(\state{}\right) \coloneqq \frac{1}{2}\sum_{k}\mathbf{e}_{p,k}\left(\state{}\right)^{T}\mathbf{Q}_{k}\left(t_{k+1}\right)^{-1}\mathbf{e}_{p,k}\left(\state{}\right),
\end{equation*}
where the error penalizes deviation from the constant-velocity prior defined in \eqref{eq:transition},
\begin{equation}
\mathbf{e}_{p,k}\left(\state{}\right) = \boldsymbol{\gamma}_{k}\left(t_{k+1}\right) -  \boldsymbol{\Phi}\left(t_{k+1},t_k\right)\boldsymbol{\gamma}_{k}\left(t_k\right),
\label{eq:priorerror}
\end{equation}
and the covariance block matrix is defined in terms of the power spectral density, $\mathbf{Q}_c$, of the local white noise process,
\begin{equation*}
\mathbf{Q}_{k}\left(t\right) \coloneqq \begin{bmatrix}
\frac{1}{3}\left(t - t_k\right)^{3}\mathbf{Q}_c & \frac{1}{2}\left(t - t_k\right)^{2}\mathbf{Q}_c \\
\frac{1}{2}\left(t - t_k\right)^{2}\mathbf{Q}_c & \left(t - t_k\right)\mathbf{Q}_c
\end{bmatrix}.
\end{equation*}

The total cost, $J\left(\state{}\right)$, is minimized by linearizing the error about an operating point, $\state{}_{\mathrm{op}}$. 
The operating point is perturbed according to the transform perturbations, $\{\boldsymbol{\epsilon}_k \in \mathbb{R}^{6}\}$, velocity perturbations $\{\boldsymbol{\psi}_k \in \mathbb{R}^{6}\}$, and landmark perturbations, $\{\boldsymbol{\zeta}_j \in \mathbb{R}^{3}\}$, which are stacked to form the full state perturbation, $\delta\state{}$. \looseness=-1

Linearizing the cost function requires linearizing  \eqref{eq:meas_error} and \eqref{eq:priorerror}.
Using the Jacobians of the measurement error function, $\mathbf{G}_{jk}$, and the prior error function, $\mathbf{E}_{k}$, the linearized cost is given by\looseness=-1
\begin{equation}
J\left(\state\right) \approx J(\state{}_{\mathrm{op}}) - \mathbf{b}^T\delta\state{} +\frac{1}{2}\delta\state{}^T\mathbf{A}\delta\state{},
\label{eq:linear_cost} 
\end{equation}
where,
\begin{equation*}
\begin{aligned}
\mathbf{b} &= \sum_{jk}\mathbf{P}_{jk}^{T}\mathbf{G}_{jk}^{T}\mathbf{R}_{jk}^{-1}\mathbf{e}_{y,jk}\left(\state{}_\mathrm{op}\right) + \sum_{k}\mathbf{P}_{k}^{T}\mathbf{E}_{k}^{T}\mathbf{Q}_{k}^{-1}\mathbf{e}_{p,k}\left(\state{}_\mathrm{op}\right),\\
\mathbf{A} &=  \sum_{jk}\mathbf{P}_{jk}^{T}\mathbf{G}_{jk}^{T}\mathbf{R}_{jk}^{-1}\mathbf{G}_{jk}\mathbf{P}_{jk} +  \sum_{k}\mathbf{P}_{k}^{T}\mathbf{E}_{k}^{T}\mathbf{Q}_{k}^{-1}\mathbf{E}_{k}\mathbf{P}_{k}.
\end{aligned}
\end{equation*}
The indicator matrices $\mathbf{P}_{jk}$ and $\mathbf{P}_k$ are defined such that $\delta\state{}_{jk} = \mathbf{P}_{jk}\delta\state{}$ and $\delta\state{}_{k} = \mathbf{P}_{k}\delta\state{}$, respectively.

The Jacobian of the measurement function is given by 
\begin{equation}
\mathbf{G}_{jk} \coloneqq \frac{\partial\mathbf{g}}{\partial\state{}}\bigg\rvert_{\state{}_{\mathrm{op},jk}} = \frac{\partial\mathbf{s}}{\partial\mathbf{z}}\bigg\rvert_{\mathbf{z}(\mathbf{x}_{\mathrm{op},jk})}\frac{\partial\mathbf{z}}{\partial\state{}}\bigg\rvert_{\state{}_{\mathrm{op},jk}},\label{eq:measurementjacobian}
\end{equation}
where
\begin{equation*}
\nonumber\frac{\partial\mathbf{z}}{\partial\state{}}\bigg\rvert_{\state{}_{\mathrm{op},jk}} = \begin{bmatrix}
\left({\transform[{\mathrm{op},{C_k}{C_1}}]}\mathbf{p}_{\mathrm{op},C_1}^{{j_1}{C_1}}\right)^{\odot} & \mathbf{0} & {\transform[{\mathrm{op},{C_k}{C_1}}]}\mathbf{D}
\end{bmatrix},
\end{equation*}
$\mathbf{D} = \begin{bmatrix}\mathbf{1} & \mathbf{0}\end{bmatrix}^T$, and  $\left(\cdot\right)^{\odot}$ is defined in \cite{barfoot2017}.

The Jacobian of the prior error function is
\begin{equation*}
\mathbf{E}_{k}\hspace{-0.75mm}=\hspace{-0.75mm}
\left[\begin{matrix}
\boldsymbol{\mathcal{J}}^{\scriptscriptstyle-1}_{{\scriptscriptstyle k+1,k}}\bar{\boldsymbol{\mathcal{T}}}_{{\scriptscriptstyle k+1,k}} & \hspace{-3mm}\Delta t_{\scriptscriptstyle k}\mathbf{1} & \scalebox{0.5}[1.0]{\( - \)}\boldsymbol{\mathcal{J}}^{\scriptscriptstyle-1}_{{\scriptscriptstyle k+1,k}} & \hspace{-1mm}\mathbf{0}\\
\frac{1}{2}\boldsymbol{\varpi}_{{\scriptscriptstyle k+1}}^{\curlywedge}\boldsymbol{\mathcal{J}}^{\scriptscriptstyle-1}_{{\scriptscriptstyle k+1,k}}\bar{\boldsymbol{\mathcal{T}}}_{{\scriptscriptstyle k+1,k}} & \hspace{-2mm}\mathbf{1} &\hspace{-2.25mm} \scalebox{0.5}[1.0]{\( - \)}\frac{1}{2}\boldsymbol{\varpi}_{{\scriptscriptstyle k+1}}^{\curlywedge}\boldsymbol{\mathcal{J}}^{\scriptscriptstyle-1}_{{\scriptscriptstyle k+1,k}} &\hspace{-1.75mm} \scalebox{0.5}[1.0]{\( - \)}\boldsymbol{\mathcal{J}}^{\scriptscriptstyle-1}_{{\scriptscriptstyle k+1,k}}
\end{matrix}\right],
\end{equation*}
where $\boldsymbol{\mathcal{J}}^{-1}_{k+1,k} \coloneqq \boldsymbol{\mathcal{J}}\left(\ln\left(\transform[C_{k+1}C_{k}]\right)^{\vee}\right)^{-1}$, $\Delta t_k \coloneqq t_{k+1} - t_{k}$, and $\left(\cdot\right)^{\curlywedge}$ and $\bar{\boldsymbol{\mathcal{T}}}_{k+1,k} \in \mathbb{R}^{6\times6}$, the adjoint of $\transform[C_{k+1}C_{k}]$, are defined in \cite{barfoot2017}.

The optimal perturbation, $\delta\state{}^*$, to minimize the linearized cost, $J\left(\state{}\right)$, is the solution to
$\mathbf{A}\delta\state{}^* = \mathbf{b}$. 
Each element of the operating point is then updated using 
\begin{equation*}
\begin{aligned}
{\transform[{\mathrm{op},{C_k}{C_1}}]} &\leftarrow\exp({\boldsymbol{\epsilon}_{k}^{*}}^{\wedge}){\transform[{\mathrm{op},{C_k}{C_1}}]},\\ 
{\boldsymbol{\varpi}_{C_k}} &\leftarrow {\boldsymbol{\varpi}_{C_k} + \boldsymbol{\psi}^{*}_{k}},\\ 
\mathbf{p}_{\mathrm{op},C_1}^{{j_1}{C_1}} &\leftarrow \mathbf{p}_{\mathrm{op},C_1}^{{j_1}{C_1}} + \mathbf{D}\boldsymbol{\zeta}^{*}_{j},
\end{aligned}
\end{equation*}
and the cost is relinearized about the updated operating point.
The process iterates until the state convergences and ${\state{}} \gets \state{}_\mathrm{op}$. 

\subsubsection{Third-Party Estimation}\label{sec:methodology:geocentric}
The geocentric motions~not belonging to the camera, $\left\{\disctraj_\mylabel{}\right\}_{\mylabel\in\mathcal{L}\setminus C}$, are calculated using the estimated egomotion, $\disctraj_C$.
As in Section \ref{sec:methodology:egocentric}, each label's state, $\state{}$, comprises its pose and velocity estimates, $\left\{\disctraj{}_{\mylabel{}_k}\right\}_{k=1,\dots,K}$, and its associated landmark points, $\left\{\mathbf{p}_{C_{1}}^{{j_{1}}{C_{1}}}\right\}_{j=1,\dots,\vert\mathcal{P}_\mylabel{}\vert}$.

The transform model, $\mathbf{z}$, used by \cite{anderson2015} is invalid for third-party motions and must be 
adjusted to transform egocentrically observed points through a geocentrically estimated state,\looseness=-1
\begin{equation*}
\mathbf{z}'\left(\mathbf{x}_{jk}\right) \coloneqq\transform[{{C}_{k}{C}_{1}}]\transform[{\mylabel{}_{1}{C}_{1}}]^{-1}\transform[{\mylabel{}_{k}\mylabel{}_{1}}]^{-1}\mathbf{F}_{\mylabel{}_{k}\mylabel{}_{1}}\transform[{\mylabel{}_{1}C_{1}}]\mathbf{p}_{C_{1}}^{{j_{1}}{C_{1}}},
\end{equation*}
where $\mathbf{F}_{\mylabel{}_{k}\mylabel{}_{1}}$ is the object deformation matrix (identity for rigid bodies), and $\transform[{{C}_{k}{C}_{1}}]$ is the camera egomotion as estimated in Section \ref{sec:methodology:egocentric}.
The transform from the camera to the object centroid is given by 
\begin{equation}
\transform[{\mylabel{}_{1}C_{1}}] = 
\begin{bmatrix}
\mathbf{C}_{\mylabel_{1} C_{1}} & \mathbf{r}_{\mylabel{}_{1}}^{C_{1}\mylabel{}_{1}}\\
\mathbf{0}^T & 1
\end{bmatrix}=\begin{bmatrix}
\mathbf{C}_{\mylabel_{1} C_{1}} & \hspace{-1.5mm}-\mathbf{C}_{\mylabel_{1} C_{1}}\mathbf{r}_{C_{1}}^{\mylabel{}_{1}C_{1}}\\
\mathbf{0}^T & 1
\end{bmatrix}.
\label{eq:initialcentroid}
\end{equation}
The rotation, $\mathbf{C}_{\mylabel_{1} C_{1}}$, is arbitrary and initially assumed to be identity for new motions.
The translation, $\mathbf{r}_{C_{1}}^{\mylabel{}_{1}C_{1}}$
, is calculated as the centroid of the labeled points, $p^{j} \in \mathcal{P}_\mylabel$, observed in the first frame. 
In a sliding-window pipeline, this transform can be determined from the previously estimated trajectory estimates.\looseness=-1


The motion model part of the measurement Jacobian is given by the block-row vector,
\begin{equation*}
\begin{split}
\frac{\partial\mathbf{z}'}{\partial\state{}}\bigg\rvert_{\state{}_{\mathrm{op},jk}} =\left[-\transform[C_{k}C_{1}]\transform[\mylabel{}_{1}C_{1}]^{-1}\transform[\mathrm{op},\mylabel{}_{k}\mylabel{}_{1}]^{-1}\left(\mathbf{F}_{\mylabel{}_{k}\mylabel{}_{1}}\transform[\mylabel{}_{1}C_{1}]\mathbf{p}_{\mathrm{op},C_{1}}^{{j_{1}}{C_{1}}}\right)^{\odot} \right.\\
\left.\begin{matrix}
 \mathbf{0}  & \transform[C_{k}C_{1}]\transform[\mylabel{}_{1}C_{1}]^{-1}\transform[\mathrm{op},\mylabel{}_{k}\mylabel{}_{1}]^{-1}\mathbf{F}_{{\mylabel{}_{k}}\mylabel{}_{1}}\transform[\mylabel{}_{1}C_{1}]\mathbf{D}
\end{matrix}\right]\hspace{1mm}
\end{split}
\end{equation*}
This Jacobian is used to estimate $\mathbf{G}_{jk}$ in  \eqref{eq:measurementjacobian}, and \eqref{eq:linear_cost} is used to estimate the continuous-time geocentric trajectory, $\disctraj{}_{\mylabel}$, of every third-party motion in the scene.

\subsection{Trajectory Extrapolation} \label{sec:methodology:extrapolation}
Motion priors are used to extrapolate motions in the presence of occlusions.
The local state, $\boldsymbol{\gamma}_k$, at time $t_k$ can be used in \eqref{eq:transition} to estimate the extrapolated state, $\hat{\boldsymbol{\gamma}}_k$, at time $\tau$ \cite{anderson2015}.
The extrapolated state is then transformed to the global state, consisting of the extrapolated transform, $\transform[]{}_{\hat{\mylabel}}$, and velocity, $\boldsymbol{\varpi}_{\hat{\mylabel}}$, via\looseness=-1
\begin{equation*}
\begin{aligned}
\transform[]{}_{\hat{\mylabel}}\left(\tau\right) &= \exp\left(\begin{bmatrix}
\mathbf{1} &  \mathbf{0}\end{bmatrix}\hat{\boldsymbol{\gamma}}_{k}\left(\tau\right)\right){\transform[\mylabel]{}}\left(t_k\right),\\
\boldsymbol{\varpi}_{\hat{\mylabel}}\left(\tau\right) &= \begin{bmatrix}
\mathbf{0} & \mathbf{1}\end{bmatrix}\hat{\boldsymbol{\gamma}}_{k}\left(\tau\right).
\end{aligned}
\label{eq:extrapolate}
\end{equation*}

Estimates can be extrapolated forward or backward in time.
Their accuracy depends on the fidelity of the motion prior to the true object motions, i.e., if their velocity is near-constant.
As the length of the extrapolation grows, the estimates will drift from the true motion of the object, especially if it exhibits significant changes in velocity.
If the extrapolation error is large, the trajectory will be treated as a newly discovered motion when it becomes unoccluded.

\subsection{Motion Closure and Interpolation} \label{sec:methodology:motionclosure}
The extrapolated estimates of previously seen motions can be used to identify unoccluded motions through \emph{motion closure}.
If a newly discovered motion is sufficiently similar to an occluded motion, then it is taken to be the reappearance of that previous motion.

Each newly discovered trajectory, $\disctraj_{\mylabel{}_k'} \coloneqq \left\{\transform[\mylabel{}'_{k}\mylabel{}'_{1}], \boldsymbol{\varpi}_{\mylabel{}'_k} \right\} $, is compared to an occluded motion's extrapolated trajectory at time $t_k$ using both position and velocity \cite{juddthesis}.
If this motion-based similarity is less than a user-defined threshold, the motions are closed; otherwise, the new trajectory is considered a new motion.
The corrected trajectory, $\disctraj_{\mylabel{}_{k}} \coloneqq \left\{\transform[\mylabel{}_{k}\mylabel{}_{1}], \boldsymbol{\varpi}_{\mylabel{}_k} \right\} $
, is then estimated from the extrapolated trajectory, $\disctraj_{\hat{\mylabel{}}_{k}}\coloneqq \left\{\transform[\hat{\mylabel{}}_{k}\hat{\mylabel{}}_{1}], \boldsymbol{\varpi}_{\hat{\mylabel{}}_k} \right\} $, and the closure transform, $\transform[\mylabel{}_{k}\hat{\mylabel{}}_{k}]$,\looseness=-1
\begin{equation*}
\nonumber\transform[\mylabel{}_{k}\mylabel{}_{1}]{}=\transform[\mylabel{}_{k}\hat{\mylabel{}}_{k}]{}\transform[\hat{\mylabel{}}_{k}\hat{\mylabel{}}_{1}]\transform[\hat{\mylabel{}}_{1}{\mylabel{}}_{1}],\qquad \boldsymbol{\varpi}_{\mylabel_k} = \boldsymbol{\varpi}_{{\mylabel'}_k},
\end{equation*}
where $\transform[\hat{\mylabel}_1\mylabel_1]{}$ is identity because the corrected and extrapolated trajectories are equivalent before the occlusion.
The corrected velocity is taken directly from the newly discovered velocity. 

\begin{figure}[t]
	\centering
	
	\includegraphics[clip,width=\columnwidth,page=1]{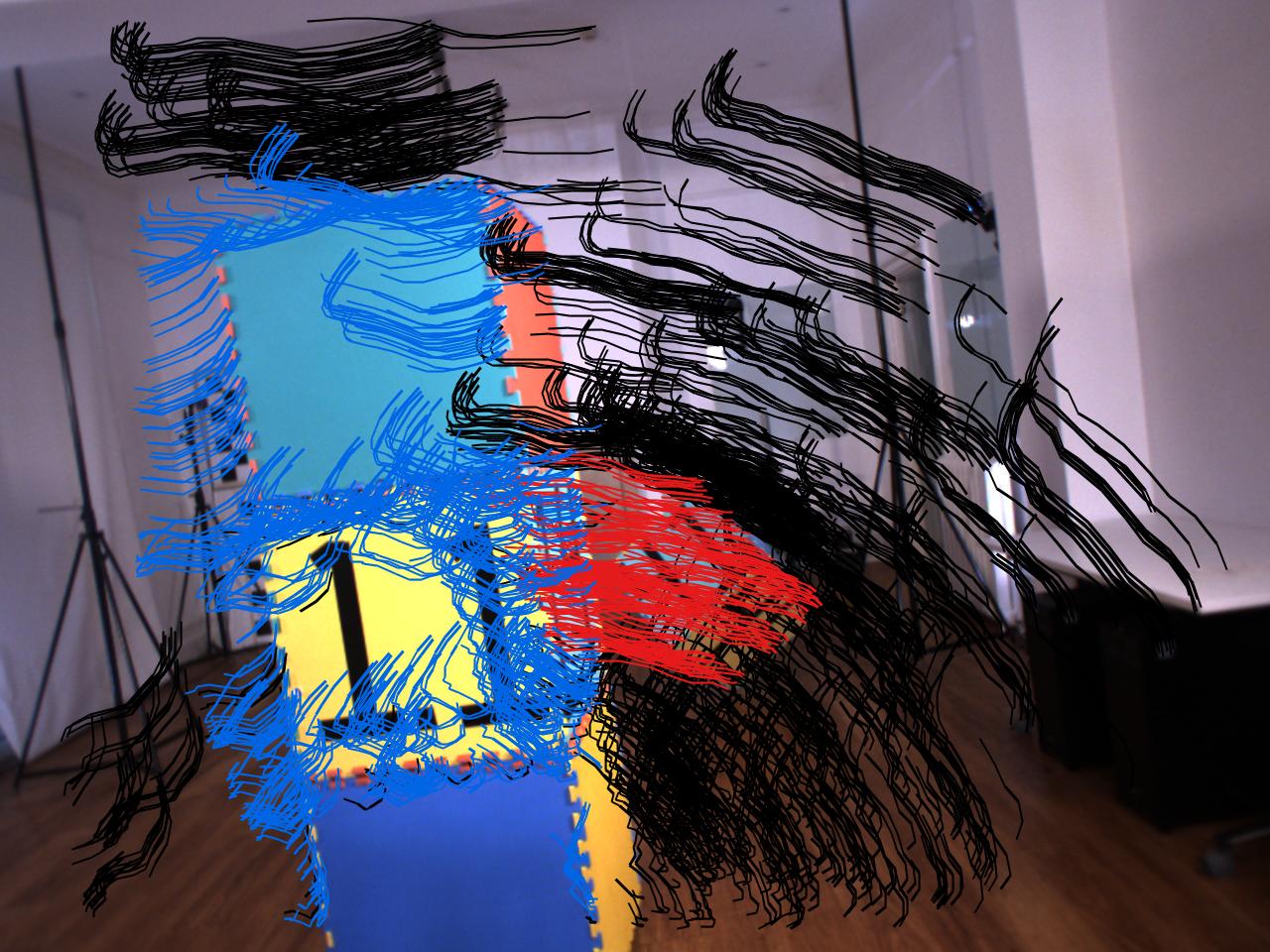}%
	
	\caption{Motion segmentation produced by occlusion-robust MVO. The egomotion of the camera is estimated from the static points in the scene shown in black. The motions of the swinging block (4, red) and the block tower (1, blue) are segmented and estimated simultaneously with the egomotion.\looseness=-1}
		\vspace{-3.5mm}
	\label{fig:segmentation}
\end{figure}

\begin{figure}[t!]
	\centering
	\subfloat[Camera Egomotion]{
		\centering
		\includegraphics[clip,width=.99\columnwidth,page=1]{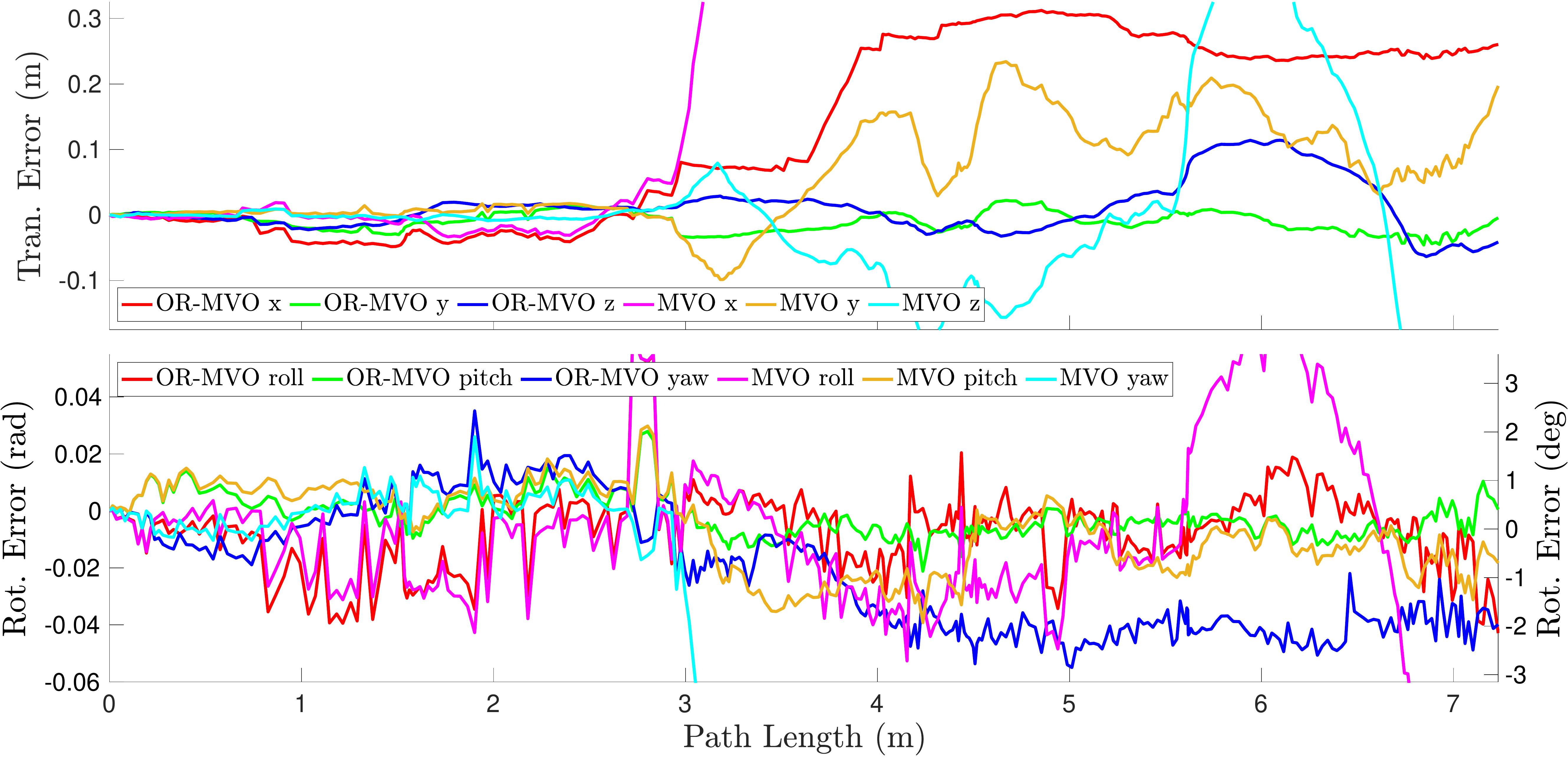}
		\label{fig:results:ego}
	}\\
	\vspace{-1mm}	
	\subfloat[Block Tower]{
		\centering
		\includegraphics[clip,width=.99\columnwidth,page=1]{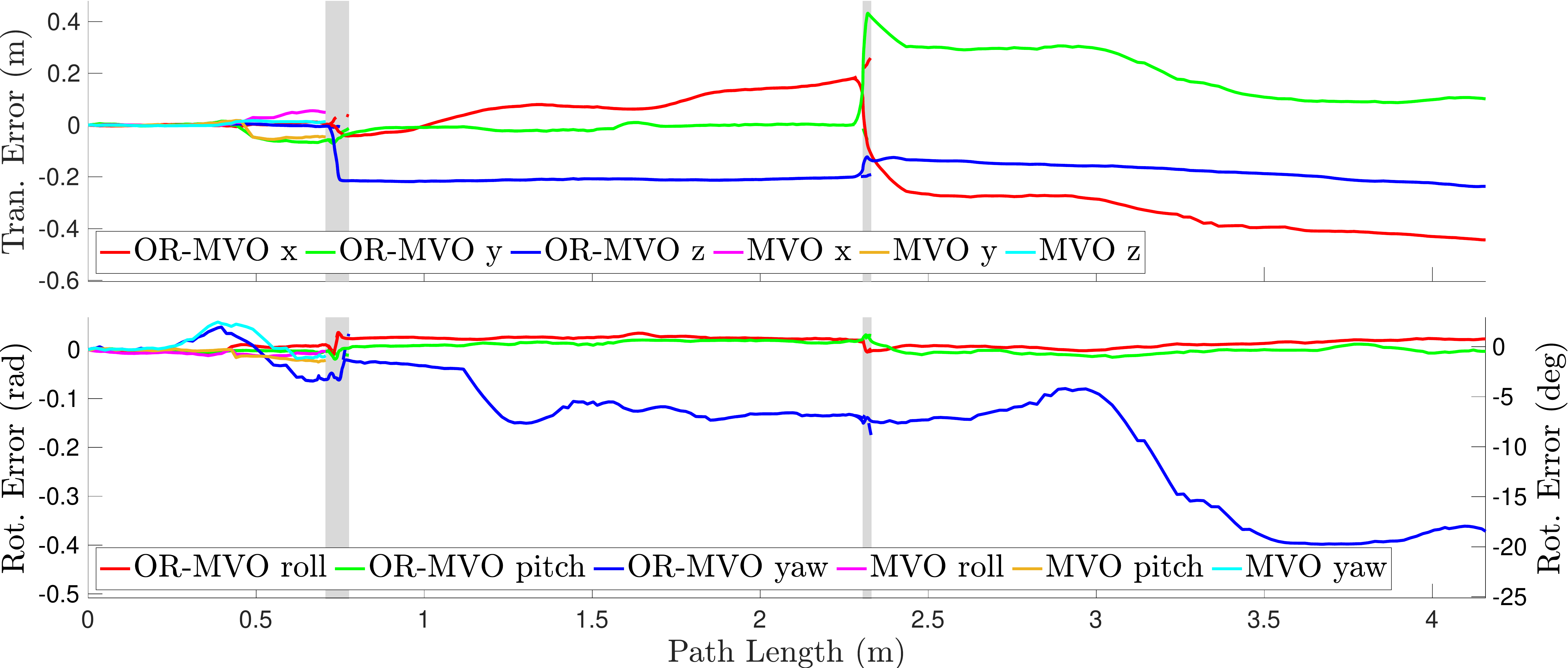}
		\label{fig:results:box1}
	}\\    
	\vspace{-1mm}	
	\subfloat[Swinging Block]{
		\centering
		\includegraphics[clip,width=.99\columnwidth,page=1]{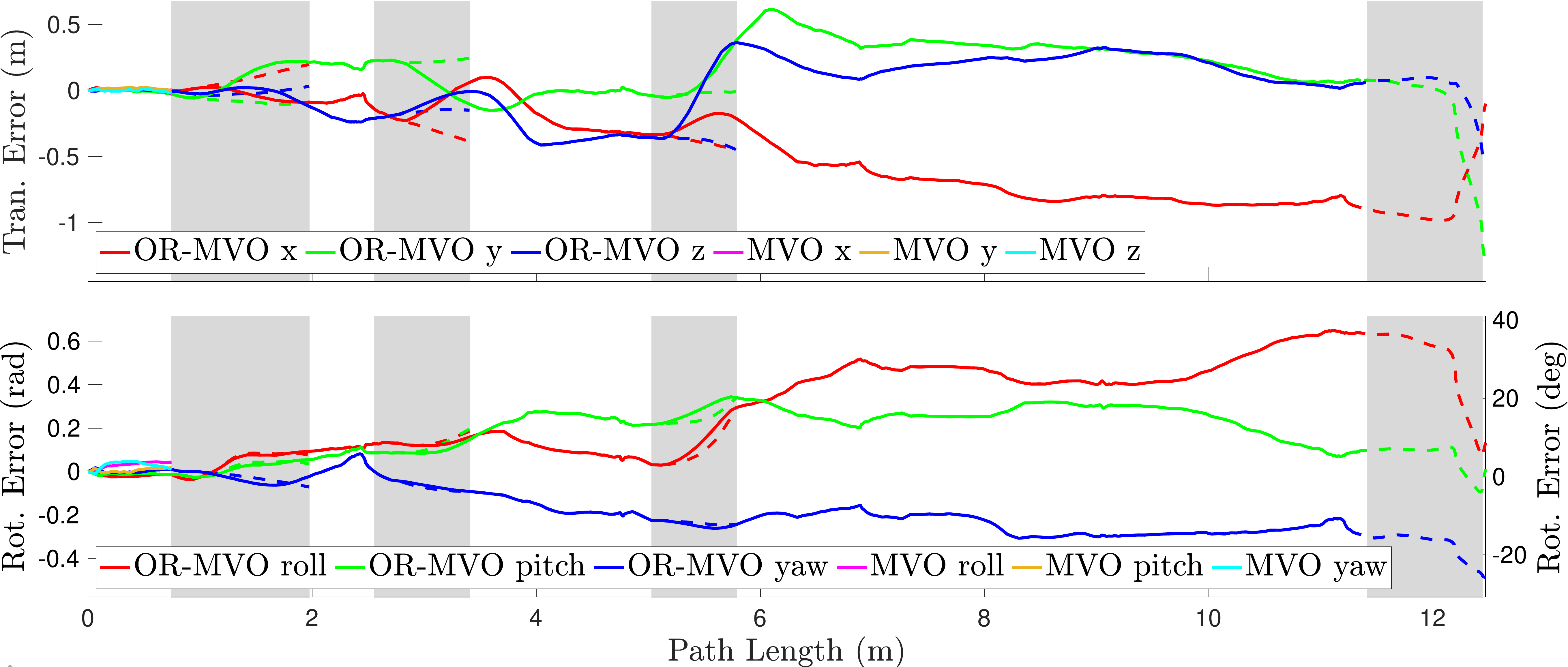}
		\label{fig:results:box4}
	}
	\vspace{0.75mm}
	\caption{The translational and rotational errors in the estimated motion of the camera (a), the block tower (b), and the swinging block (c) for a $300$-frame section of \texttt{occlusion\_2\_unconstrained} segment of the OMD.
		The occlusion-robust MVO errors are displayed in red-green-blue, and the original MVO errors are displayed in magenta-yellow-cyan.
		Each object is compared to ground-truth trajectory data and errors are reported in an arbitrary geocentric frame with the z-axis up and arbitrary x- and y-axes.
		Grey regions show when the swinging block was occluded by the tower, or when the tower was stationary and effectively part of the background.
		Dashed lines represent the extrapolation error and the solid lines represent the error in the direct or interpolated estimates.
		As expected, the original MVO pipeline failed to track the blocks past their first occlusions.
		This also led to significant errors in the egomotion estimate that have been cropped in (a).\looseness=-1}
		\vspace{-3mm}
	\label{fig:results}
\end{figure}

The closure transform,
\begin{equation*}
\transform[\mylabel{}_{k}\hat{\mylabel{}}_{k}]{}=\begin{bmatrix}
\mathbf{C}_{\mylabel{}_{k}\hat{\mylabel{}}_{k}} & \mathbf{r}^{\hat{\mylabel{}}_{k}\mylabel{}_{k}}_{\mylabel{}_{k}}\\
\mathbf{0}^T & 1
\end{bmatrix},
\end{equation*} 
shifts the extrapolated trajectory position, $\mathbf{r}^{\hat{\mylabel{}}_{k}C_{k}}_{C_{k}}$, to the observed centroid of the newly discovered trajectory, $\mathbf{r}^{\mylabel{}_{k}'C_{k}}_{C_{k}}$.

\kjtodo{The closure rotation, $\mathbf{C}_{\mylabel{}_{k}\hat{\mylabel{}}_{k}}$, and the closure translation, $\mathbf{r}^{\hat{\mylabel{}}_{k}\mylabel{}_{k}}_{\mylabel{}_{k}}$, are calculated independently.}
The closure rotation is taken as identity because it is difficult to determine the change in rotation of the object after the occlusion without using appearance-based metrics.
This carries forward the extrapolation of the original motion's rotation.

The closure translation is given as
\begin{align}
\nonumber\mathbf{r}^{\hat{\mylabel{}}_{k}\mylabel{}_{k}}_{\mylabel{}_{k}} &=\mathbf{C}_{\mylabel_{k}C_{k}}\mathbf{r}^{\hat{\mylabel{}}_{k}\mylabel{}_{k}}_{C_{k}} = \mathbf{C}_{\mylabel_{k}C_{k}}\mathbf{r}^{\hat{\mylabel{}}_{k}\mylabel{}_{k}'}_{C_{k}}\\ 
\nonumber&=\mathbf{C}_{\mylabel_{k}\hat{\mylabel}_{k}}\mathbf{C}_{\hat{\mylabel{}}_{k}C_k}\left(\mathbf{r}^{\hat{\mylabel{}}_{k}C_{k}}_{C_{k}} - \mathbf{r}^{\mylabel{}_{k}'C_{k}}_{C_{k}}\right).\\
\intertext{The extrapolated camera-object rotation, $\mathbf{C}_{\hat{\mylabel{}}_{k}C_k}$, is~taken~from}
\nonumber\transform[C_{k}\hat{\mylabel{}}_{k}]{}&=\begin{bmatrix}
\mathbf{C}_{\hat{\mylabel{}}_{k}C_{k}}^{T} & \mathbf{r}^{\hat{\mylabel{}}_{k}C_{k}}_{C_{k}}\\
\mathbf{0}^T & 1
\end{bmatrix}= \transform[C_{k}C_{1}]{}{\transform[{\mylabel{}}_{1}C_{1}]^{-1}\transform[\hat{\mylabel{}}_{1}\mylabel{}_{1}]^{-1}}{\transform[\hat{\mylabel{}}_{k}\hat{\mylabel{}}_{1}]^{-1}}.
\end{align}

The corrected pose and velocity, $\disctraj_{\mylabel{}_k}$, are then used to interpolate from the beginning of the occlusion at time $t_{j+1}$ and correct the extrapolated estimates.
Given the state transition function defined in \eqref{eq:transition}, the interpolation between $t_j$ and $t_k$ is defined as
\begin{align*}
\hat{\boldsymbol{\gamma}}_{j}\left(\tau\right) &= \boldsymbol{\Lambda}\left(\tau\right)\boldsymbol{\gamma}_j\left(t_j\right) + \boldsymbol{\Omega}\left(\tau\right){\boldsymbol{\gamma}}_{j}\left(t_{k}\right),\\
\intertext{where}
\boldsymbol{\Lambda}\left(\tau\right) &= \boldsymbol{\Phi}\left(\tau,t_j\right)-\boldsymbol{\Omega}\left(\tau\right)\boldsymbol{\Phi}\left(t_{k},t_j\right),\\
\boldsymbol{\Omega}\left(\tau\right) &= \mathbf{Q}_{j}\left(\tau\right)\boldsymbol{\Phi}\left(t_{k},\tau\right)^{T}\mathbf{Q}_{j}\left(t_k\right)^{-1},
\end{align*}
and $t_j < \tau < t_k$ \cite{anderson2015}.
This interpolation can explain the occluded motion of the object better than extrapolation because it includes direct estimates on both sides of the occlusion.\looseness=-1

\section{Experiments and Results} \label{sec:experiments}
The performance of occlusion-robust MVO is evaluated both quantitatively and qualitatively.
The OMD \cite{judd2019} was used to metrically evaluate the estimation accuracy of the pipeline using ground-truth trajectory data (\cref{fig:segmentation,fig:results}).
Although the OMD consists of controlled indoor experiments, its ground-truth trajectory data for \emph{every} motion in the scene makes it uniquely appropriate for quantitatively evaluating the accuracy of MVO.
The pipeline is also qualitatively evaluated on complex, piece-wise rigid motions in a real-world scenario for which there is no ground truth (\cref{fig:marquee}).

The results were produced from sequences of stereo camera data using a 16-frame sliding window.
Tracklets are generated using LIBVISO2 \cite{geiger2011} and the minimization was performed analytically with Ceres \cite{ceres-solver}.
The quantitative results were run on a $300$-frame segment from the \texttt{occlusion\_2\_unconstrained} segment of the OMD.
The motion closure similarity was calculated using the Euclidean distance between the extrapolated and observed transforms in the $\mathfrak{se}\left(3\right)$ tangent space \cite{juddthesis} and the threshold was set to 1.
The transforms between the ground-truth Vicon frames and our estimated frames are arbitrary, so the first $15$ poses are used to calibrate these transforms \cite{zhang2018}.
All errors are reported for geocentric trajectory estimates.


\kjtodo{The occlusion-robust MVO egomotion estimates (\cref{fig:results:ego}) exhibited a maximum total drift of $0.26$ m (over a $7.48$ m path).} 
This error ($3.48\%$) is reasonable compared to the drift in other similar, camera-only VO systems \cite{geiger2012}.
\kjtodo{The centroid estimates used in motion closure occasionally shifted significantly and caused the interpolated error of the block tower and the swinging block to be worse than the extrapolated error.}
The block tower exhibited a maximum total drift of $0.66$ m (over a $3.89$ m path), and  
the swinging block exhibited a maximum total drift of $1.58$ m (over a $14.12$ m path). 
\looseness=-1

As expected, the original MVO pipeline failed to track the motions of blocks through the occlusions.
The egomotion estimation error was also higher ($1.10$ m) because the block tower may be mistakenly labeled as the static background when it begins to move again.  
In contrast, the occlusion-robust MVO pipeline both tracks the blocks through the occlusions and estimates the egomotion more accurately because motion closure helps disambiguate between the static background and motions starting from rest.

The qualitative results were run on data collected from a real-world street scenario for which ground-truth trajectory data is prohibitively difficult to collect (\cref{fig:marquee}).
The camera pans to follow a cyclist moving from left to right across the scene as they are temporarily occluded by a truck moving from right to left.
Both objects are consistently segmented and estimated, along with the camera egomotion, despite the cyclist exhibiting some nonrigid motion.
The constant-velocity assumption proved reasonable in this scenario as the cyclist's motion was accurately extrapolated through the occlusion and motion closure was successful.

\section{Discussion} \label{sec:discussion}

The occlusion-robust pipeline consistently segments the motions of the camera, blocks, cyclist, and truck while also estimating the trajectories through occlusions; however, it is still a sparse, feature-based technique.
The observable shape (and centroid) of an object changes as it moves, affecting the geocentric estimate of its trajectory.
As the object becomes more occluded, the quality of the trajectory estimation will degrade.
It is unlikely that an object will become occluded or unoccluded instantaneously, but this can be mitigated by predicting occlusions as in \cite{mitzel2010}.

The reliance of the pipeline on motion means that objects that temporarily have the same motion will be given the same label, such as when a dynamic object becomes stationary.
This is often desirable, as it implicitly handles trajectory merging, but many applications might require a form of \emph{motion permanence}.
This could be introduced by using appearance-based techniques or by explicitly modeling trajectory merging.\looseness=-1

The accuracy of the object centroid, and therefore the trajectory, depends on the distribution of observed features.
The newly calculated centroid of a motion after an occlusion does not always match the original centroid, especially if it is still partially occluded.
This discrepancy can cause jumps in the trajectory and is a major source of error in the estimates of the block tower (\cref{fig:results:box1}) because it is often partially outside the view of the camera.  
The original MVO pipeline partially mitigated this through a rolling-average calculation of the centroid, but this is difficult in geocentric estimation as the centroid is required for the calculation of third-party trajectories.
More robust centroid calculation remains an area of ongoing work, including considering part-to-whole extrapolation techniques \cite{wu2007,hu2012,shu2012}.

The qualitative results (\cref{fig:marquee}) illustrate that occlusion-robust MVO is readily applicable to real-world scenarios.
The pipeline is able to segment and estimate the motions of both the cyclist and the truck simultaneously with the egomotion of the camera.
When the cyclist is occluded, their trajectory is extrapolated using the constant-velocity prior until they are unoccluded and motion closure is successful.

\kjtodo{Piecewise-rigid motions consist of a number of small rigid motions. 
If the motions are largely consistent, MVO groups them into a bulk motion; otherwise, they are rejected as noise.}
Though the cyclist in \cref{fig:marquee} exhibits some piece-wise rigid or nonrigid motion, the overall motion is largely rigid and the robustness of MVO means the trajectory is reliably estimated.

The applicability of the white-noise-on-acceleration prior is limited in scenes where objects change direction or speed.
Few objects move with truly constant velocity, but these results illustrate that it is robust to small changes in velocity and allows velocities to evolve over time.
Future work will consider a white-noise-on-jerk prior \cite{tang2019} to better model smoothly varying accelerations, and more expressive priors can be used to address complex dynamic motions, such as when the swinging block changed direction at the end of the segment in \cref{fig:results:box4}.\looseness=-1

\section{Conclusion} \label{sec:conclusion}
This paper extends the MVO pipeline to complex, dynamic environments with significant occlusions. 
Occclusion-robust MVO uses an \se{} motion prior to metrically estimate both camera egomotion and third-party motions even in the presence of temporary occlusions.
It uses motion closure to recognize the reappearance of previously observed motions and improve estimates of their unobserved trajectories. 
Its performance~is evaluated on a challenging segment from the Oxford Multimotion Dataset (OMD) \cite{judd2019} exhibiting significant occlusion and complex \se{} motions, as well as a real-world scenario with piecewise-rigid motion.
We are currently exploring extensions to the pipeline as discussed in Section \ref{sec:discussion}, as well as~its applicability to other sensors, such as RGB-D and event cameras.\looseness=-1

\end{document}